\documentclass{bmvc2k}

\usepackage{times}
\usepackage{epsfig}
\usepackage{graphicx}
\usepackage{amsmath}
\usepackage{dsfont}
\usepackage{amssymb}
\usepackage{mathrsfs}
\usepackage{tabularx}
\usepackage{makecell}
\usepackage{bm}
\usepackage{hhline}
\usepackage[dvipsnames]{xcolor}
\mathchardef\mhyphen="2D 

\newcommand\NOTES[1]{\textbf{\textcolor{blue}{}} }
\newcommand{\note}[1]{{\bf Note:} \{\uline{}\}}
\newcommand{\fixme}[1]{{\Large FIXME:} {\bf }}
\newcommand{\todo}[1]{{\color{red}\bf TODO: \uline{}}}
\newcommand{\TODO}[1]{{\color{red}\bf TODO: \uline{}}}

\title{Adaptive Appearance Rendering}

\addauthor{Mengyao Zhai}{mzhai@sfu.ca}{1}
\addauthor{Ruizhi Deng}{ruizhid@sfu.ca}{1}
\addauthor{Jiacheng Chen}{jca348@sfu.ca}{1}
\addauthor{Lei Chen}{chenleic@sfu.ca}{1}
\addauthor{Zhiwei Deng}{zhiweid@sfu.ca}{1}
\addauthor{Greg Mori}{mori@cs.sfu.ca}{1}

\addinstitution{
 School of Computing Science\\
 Simon Fraser University\\
 Burnaby, Canada
}

\runninghead{Zhai et al.}{Adaptive Appearance Rendering}

\begin{document}

\maketitle

\begin{abstract}

We propose an approach to generate images of people given a desired appearance and pose. Disentangled representations of pose and appearance are necessary to handle the compound variability in the resulting generated images. Hence, we develop an approach based on intermediate representations of poses and appearance: our pose-guided appearance rendering network firstly encodes the targets' poses using an encoder-decoder neural network. Then the targets' appearances are encoded by learning adaptive appearance filters using a fully convolutional network. Finally, these filters are placed in the encoder-decoder neural networks to complete the rendering. We demonstrate that our model can generate images and videos that are superior to state-of-the-art methods, and can handle pose guided appearance rendering in both image and video generation.

\end{abstract}


\begin{figure}[h]
\center
\includegraphics[width=0.95\textwidth]{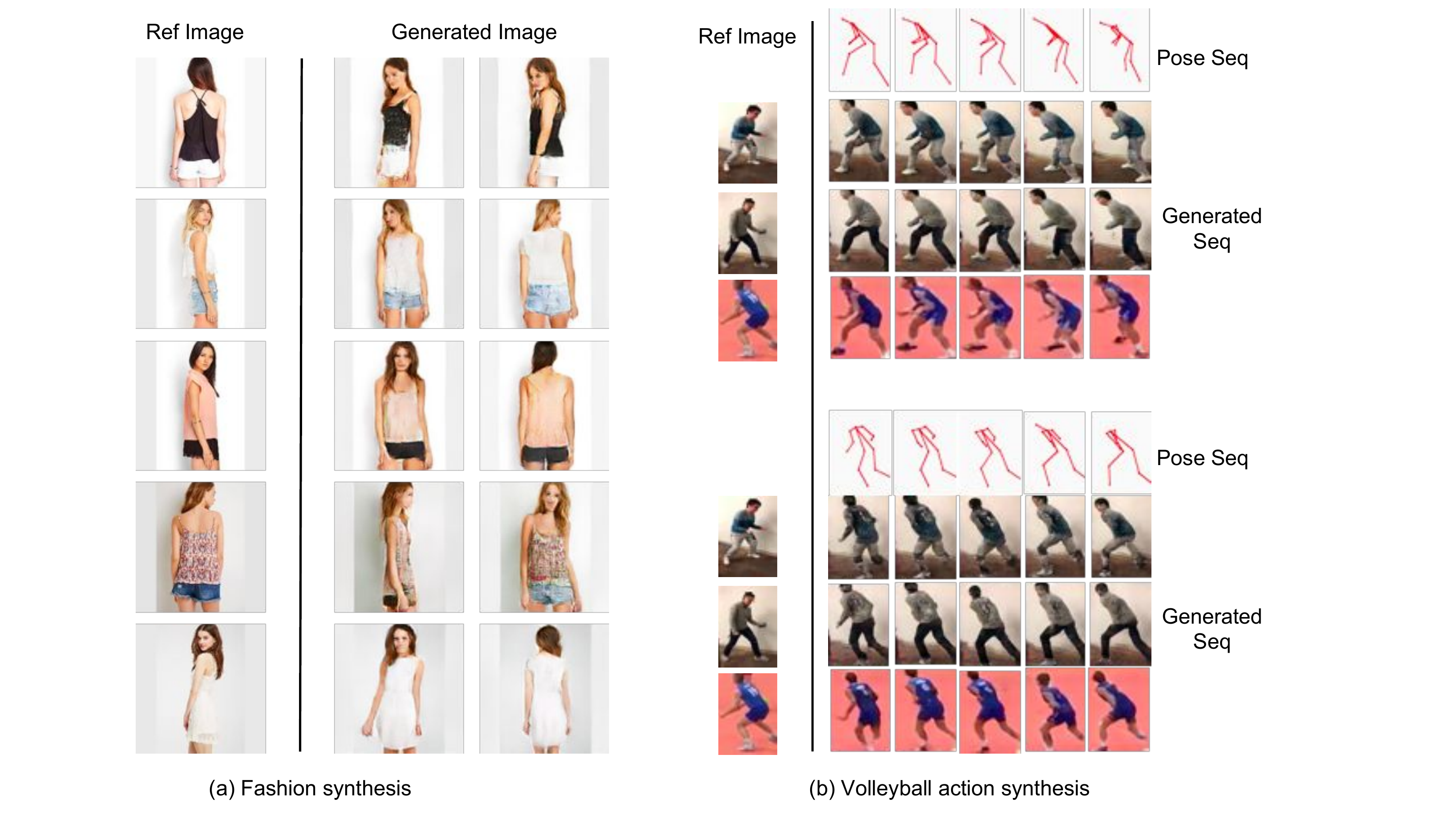}
\caption{We develop architectures for forecasting person images of various appearances.  At the core of the method is adaptive rendering modules. A series of generation examples of various reference images rendered into new poses is shown above.}
\label{fig-intro}
\end{figure}


\section{Introduction}

We may not be able to play soccer like Lionel Messi, but perhaps we can train deep networks to hallucinate imagery suggesting that we can.  Consider the images in Fig.~\ref{fig-intro}.  In this paper we describe research toward synthesizing realistic images or sequences that forecast the appearance of people performing desired actions.  The model can take a pose and a reference image of a person, and generate a novel view of the person in the given pose, while capturing fine appearance details.

Image or video generation has emerged as an important problem in many domains. Within robotics, work has explored predicting the consequences after interactions between an agent and its environment~\cite{finn2016unsupervisedPhysicalInteraction}. In natural language processing, approaches~\cite{mansimov2015generating,reed2016generative} have been proposed to tackle tasks such as text to image or image to text synthesis. Generative models of video sequences~\cite{walker2017poseKnow, Villegas2017LearningToGenerate} and fashion images~\cite{ZhaoWCLF17,MaDPIG17} are a core part of visual understanding and have received renewed attention from the vision community.

In this paper, we focus on generating realistic images of a particular person striking a desired pose. Human body pose is a natural intermediate representation for this generation, and hence utilized in many previous methods for synthesizing human motion and video~\cite{BrandH00,EfrosBMM03,walker2017poseKnow}. We follow in this paradigm, using body poses to generate person images and videos. Simple networks~\cite{walker2017poseKnow} may generate blurry and distorted images.  Stylistic methods~\cite{isola2016pix2pix} have shown great success in generating realistic images, but lack control over the appearance of the generated images. Our task requires the model to be able to generate images of a person with a specific appearance. Inspired by~\cite{chen2017stylebank}, we propose a novel appearance rendering network which encodes appearance into convolutional filters. These filters are operationalized using a fully convolutional network, and utilized in an image-to-image translation structure that transfers the desired appearance to the generated image.

To sum up, we contribute a new state of the art generative model that performs adaptive appearance rendering to create accurate depictions of human figures in these human poses. We demonstrate our model on two applications: fashion image synthesis and multi-person video synthesis involving complex human activities. Comprehensive quantitative results are provided to facilitate the analysis of each component of our model, and qualitative visualizations are shown in complementary to quantitative results.

\section{Related Work}

{\bf Image generation:} Image-to-image translation has achieved great success since the emergence of GANs~\cite{goodfellow2014GAN}. Recent work produces promising results using GAN-based models~\cite{isola2016pix2pix, CycleGAN2017}. Stylized images can be generated by using feed-forward networks~\cite{gatys2015neuralStyleTransfer} with the help of perceptual loss~\cite{johnson2016perceptualLoss}. The recent work of \cite{chen2017stylebank} proposes a structure to disentangle style and content for style transfer. Styles are encoded using a stylebank (set of convolution filters). Visual analogy making~\cite{NIPS2015_5845_analogy, sadeghi2015visalogy} generates or searches for an new image analagous to an input one, based on other previously given example pairs.

{\bf Generation from intermediate representations:} Generation directly in low-level pixel space is difficult and these types of approaches tend to generate blurry or distorted future frames. To tackle this problem, hierarchical models~\cite{walker2017poseKnow, Villegas2017LearningToGenerate, ZhuPrada17, KaracanAEE16, ma2017pose} adopt intermediate representations. This type of approach can alleviate image blur, however the quality of generation largely depends on the the image generation network. Simple generation networks can still produce blurry images as shown in~\cite{walker2017poseKnow}. It's worth noting that Ma et al. propose a two-stage approach to solve a problem similar to our work~\cite{ma2017pose}. However, the main contribution of this work is to propose a simple yet well-performing base model to the problem of generating images conditioned on pose and appearance.

{\bf Video generation:} Data-driven video generation has seen a renaissance in recent years. One major branch of methods uses RNN-based models such as encoder-decoder LSTMs for direct pixel-level video prediction~\cite{ranzato2014videoGenerationBaseline, srivastava2015unsupervisedVideoRepresentationLSTMs, oh2015actionConditionedVideoPrediction, mathieu2015deepBeyondMSE}. These methods successfully synthesized low-resolution videos with relatively simple semantics, such as moving MNIST digits or human action videos with very regular, smooth motion. Subsequent work has attempted to expand the quality of predicted video in terms of resolution and diversity in human activity.  Earlier efforts were focused on optical flow-timescale prediction, further work pushed past into more complex motions (e.g.~\cite{liu2017video}).

In summary, our approach builds on the substantial body of related work in pose analysis and style/analogy-based image generation. We contribute a novel, simple and effective method for adaptive appearance rendering model for image/video generation from human poses.

\begin{figure*}[!tph]
\center
\includegraphics[width=0.85\textwidth]{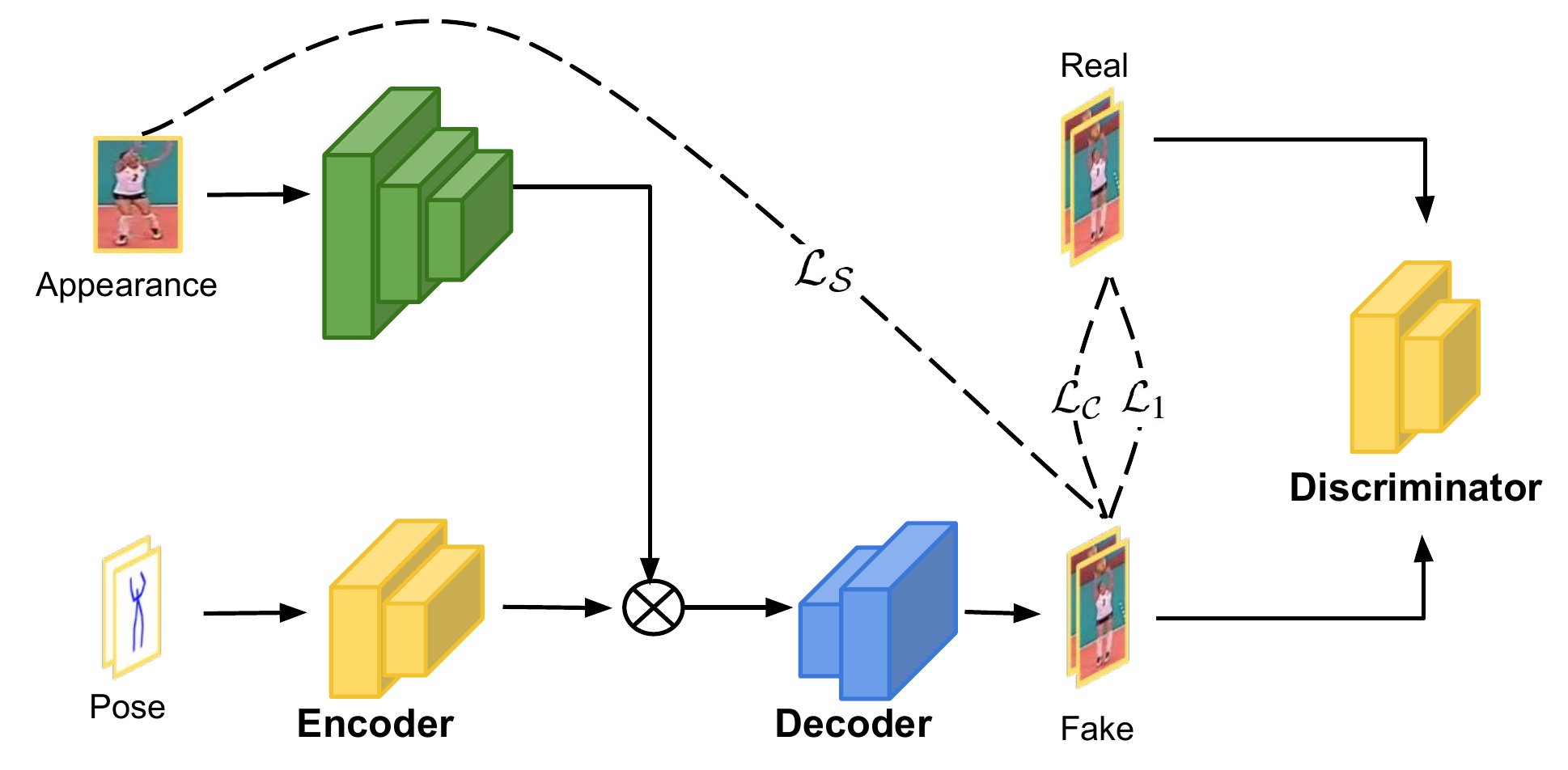}
\caption{Overview of the adaptive appearance rendering network. Given input posemap image and appearance reference image, the appearance is encoded into convolutional filters, then these filters are placed in the image-to-image translation network to transfer the appearance to images of each person striking the desired pose.}
\label{fig:overview}
\end{figure*}
%
%
\section{Adaptive Rendering Network}
\label{sec:adaptive_render}
Given pose estimations as intermediate representations, the goal of our model is to synthesize realistic image of person with desired appearance and pose, where the pose of every person is a \textit{posemap} image in which ``white" body joint points are drawn on a black background canvas. Great success has been shown in generating realistic images given a sketch as inputs~\cite{isola2016pix2pix}, which is similar to our task since the posemap image can be treated as a sketch. However, for our task we cannot generate random appearance and we need control over appearance of generated images to make sure that the generated person is wearing the desired clothing.

To accomplish this goal, we propose an adaptive rendering structure where the appearance filters are adaptively computed from an input reference image using a fully convolutional neural network (FCN) and supervised using style loss to enforce similar color distributions as the input reference image. By incorporating this FCN into an image-to-image translation network a realistic image of a target consistent with the desired action and appearance can be generated.

\subsection{Network Structure}
 
Fig.~\ref{fig:overview} shows our adaptive rendering network (Ada-R Network) architecture, which consists of two branches: an encoder-decoder branch, and an adaptive rendering branch. The network requires two input images: a posemap image, and a reference image which provides the appearance of the same person in a different pose. The goal of the network is to generate a realistic image of a person consistent with the posemap and having appearance consistent with the reference image.
 
\textbf{Encoder-Decoder}: Instead of training an encoder-decoder network which can reconstruct input images, our encoder-decoder branch shown in Fig.~\ref{fig:overview} is a sketch $\rightarrow$ image model. 

We use the same input size and encoder-decoder structure as in~\cite{isola2016pix2pix}: both generator and discriminator use modules of the form convolution-BatchNorm-Relu~\cite{ioffe2015batch}, the encoder consists of convolutional layers with stride 2 and symmetrically the decoder consists of convolutional layers with fractional stride $\frac{1}{2}$. 

\textbf{Adaptive Rendering}: The encoder-decoder network takes binary posemap images as inputs which do not contain any information about the appearance of the person.  Hence, we propose to use another network to learn appearance information. By combining these two networks together we are able to generate realistic images of a person wearing the desired clothing. Here we introduce our Ada-R network. 

To transfer the desired appearance to the encoder-decoder branch, we replace the last convolutional filter in the encoder-decoder branch with our adaptive appearance transfer filter. The adaptive appearance filter $\bm{K}_{ada \mhyphen app}$ encoding appearance information of a person is derived from an input appearance reference image $\bm{I}_{app}$ using a fully-convolutional network
\begin{equation}
    \bm{K}_{ada \mhyphen app} = FCN(\bm{I}_{app})
\end{equation}
For image generation, the inputs are appearance-posemap image pairs. For video generation, the rendering of one person's posemap sequence only requires one reference image, and the frames could be obtained by performing adaptive appearance rendering frame by frame. The filter is applied to the rendering procedure by
\begin{eqnarray}
    \bm{F} &=& \mathcal{E} (\bm{I}_{pose}) \\
    \bar{\bm{F}} &=& \bm{F} \ast \bm{K}_{ada \mhyphen app}
\end{eqnarray}
where $\mathcal{E}$ is the encoder network, $\bm{I}_{pose}$ is the posemap image and $\ast$ is a regular 2D convolution operator. $\bm{F}$ is the feature map of size $w\times h\times C_{in}$ generated by the encoder network. We set the size of $\bm{K}_{ada \mhyphen app}$ to $k\times k\times C_{in}\times C_{out}$ to be compatible with $\bm{F}$, where $k$, $C_{in}$ and $C_{out}$ denote the kernel size, input channel number and output channel number of this adaptive convolution operation. $\bar{\bm{F}}$ is the feature map after applying the adaptive appearance filter to the feature map $\bm{F}$. The person $\bm{I}_{gen}$ with desired appearance is finally produced with
\begin{equation}
     \bm{I}_{gen} = \mathcal{D} (\bar{\bm{F}})
\end{equation}
where $\mathcal{D}$ is the decoder network.

We release a reference implementation\footnote{\url{https://github.com/wisdomdeng/AdaptiveRendering}} to provide the full details on the model architecture and parameter settings.

\subsection{Loss Function}

Our network is trained in an adversarial setting, where the Ada-R network is the generator $G$, and a discriminator $D$ is introduced to discriminate between real and generated images. The overall architecture is a conditional generative adversarial network conditioned on the reference image. Let $\bm{I}_{goal}$ be the target image that we try to produce, and $\bm{I}_{gen}$ be the image that the Ada-R network generated. The loss of our Ada-R network is defined as
\begin{equation}
\mathcal{L_{CGAN}}(G, D) + \mathcal{L_{T}}
\end{equation}
where $\mathcal{L_{CGAN}}(G, D)$ is the standard adversarial loss. 
$\mathcal{L_{T}}$ is our appearance transfer loss, defined below.
\begin{equation}
\mathcal{L_{T}} = \alpha \mathcal{L}_{1}(\bm{I}_{gen}, \bm{I}_{goal}) + \beta \mathcal{L_{C}}(\bm{I}_{gen}, \bm{I}_{goal}) + \gamma \mathcal{L_{S}}(\bm{I}_{gen}, \bm{I}_{app}).
\end{equation}
$\mathcal{L}_{1}$ is an $L_1$ loss that encourages pixel-level agreement between the generated image and the target image, defined as:
\begin{equation}
\mathcal{L}_{1}(\bm{I}_{gen}, \bm{I}_{goal}) = ||\bm{I}_{gen}-\bm{I}_{goal}||.
\end{equation}
$\mathcal{L_{C}}$ and $\mathcal{L_{S}}$ are the content and style loss, defined the same as Gatys et al.~\cite{gatys2015neuralStyleTransfer}.  They 
aim to preserve image structure and colour distributions respectively:
\begin{eqnarray}
\mathcal{L_{C}}(\bm{I}_{gen}, \bm{I}_{goal}) = \sum_{l \in {l_{c}}}||F_{l}(\bm{I}_{gen})-F_{l}(\bm{I}_{goal})||^2 \\
\mathcal{L_{S}}(\bm{I}_{gen}, \bm{I}_{app}) = \sum_{l \in {l_{s}}}||G_{l}(\bm{I}_{gen})-G_{l}(\bm{I}_{app})||^2
\end{eqnarray}
where $F_{l}$ is the feature map from layer $l$ of a pretrained VGG-19 network~\cite{Simonyan15VGG}. $l_{c}$ are layers of VGG-19 used to compute the content loss. $G_{l}(\cdot)$ is the Gram matrix which learns the correlations of color distribution given two input images. $l_{s}$ are layers of VGG-19 used to compute the style loss.

The final objective is defined as
\begin{equation}
G^{\star} = \arg \underset{G}{\min}\ \underset{D}{\max}\ \mathcal{L_{CGAN}}(G, D) + \mathcal{L_{T}}
\end{equation}


\section{Experiments}
\label{sec:exp}
We demonstrate our model on the DeepFashion dataset~\cite{liuLQWTcvpr16DeepFashion} and Volleyball dataset ~\cite{Ibrahim_2016_CVPR_volleyball}. For the DeepFashion dataset, the goal is to render a given appearance to different poses of same person. Due to the completeness of poses, we perform comprehensive quantitative evaluations and an ablation study on the DeepFashion dataset. We also demonstrate a novel application of our model on the Volleyball dataset where the goal is to synthesize short videos (5 frames) containing groups of people given the people in the $1^{st}$ frame as appearance reference image. 

To train our Ada-R network, we compute content loss at layer \textit{relu4-2} and style loss at layer \textit{relu1-2}, \textit{relu2-2}, \textit{relu3-2}, \textit{relu4-2} and \textit{relu5-2} of the pre-trained VGG-19 network. We set the learning rate to \textit{1e-3}, loss weights are set to bring the $L1$, content and style losses to a similar scale. For fashion dataset $\alpha=100$, $\beta=1e-4$, $\sigma=1e-14$. For Volleyball dataset, $\alpha=100$, $\beta=0.1$, $\sigma=1e-12$. To make the training stable, for DeepFashion dataset, in each iteration the generator is updated three times and the discriminator is updated one time; and for Volleyball dataset, in each iteration the generator is updated twice and the discriminator is updated one time.

We adopt Mean Square Error (MSE), Peak Signal-to-noise Ratio (PSNR), and Structural Similarity (SSIM)~\cite{wang2004image} as evaluation metrics. MSE and PSNR are pixel-wise measurements for quality of reconstructed or generated images. SSIM measures the quality of a generated image by considering the image's structural information instead of only pixel-wise errors. For both datasets, the OpenPose detector~\cite{cao2017realtime} is used to obtain corresponding poses for each target in a given image.

\subsection{Experiments on DeepFashion}
\label{eval:fashion}
DeepFashion contains fashion images of different person IDs, and most IDs contain 4 views: front, side, back and full body. We filtered out IDs without upper body view or with less than 3 views, resulting in 3418 person IDs. To train our model, we use 2395 person IDs, resulting in 14370 pose-appearance pairs. To test our model, we use 1023 person IDs, resulting in 6138 pose-appearance pairs. Original images are resized to $128 \times 128$ for both training and testing.
 

To analyze the strength of our model as well as the importance of each component in both the network structure and loss, we compare our Ada-R approach with the following approaches:

{\bf Ada-R w/o style loss}: In this baseline, we remove the style loss from the loss function.

{\bf Ada-R w/o content loss}: In this baseline, we remove the content loss from the loss function.

{\bf Ada-R w/o $L_1$ loss}: In this baseline, we remove the $L_1$ loss from the loss function.

{\bf Pose-GAN$^\star$ (PG$^\star$)}: We adopt the generation structure used by Walker et al.~\cite{walker2017poseKnow}, namely the posemap image and appearance image are concatenated as input to the image-to-image translation network and the FCN is removed. The same loss is used as our Ada-R approach.

{\bf Pose-GAN$^\star$ (PG$^\star$) w/o style loss}: We adopt the generation structure of PG and remove the style loss from the loss function.

{\bf Visual Analogy Making$^\star$ (VAM$^\star$)}: We adopt the analogy based generation structure used in~\cite{Villegas2017LearningToGenerate,NIPS2015_5845_analogy}. Same loss is used as our Ada-R approach.

\begin{figure}[h]
\center
\includegraphics[width=0.95\textwidth]{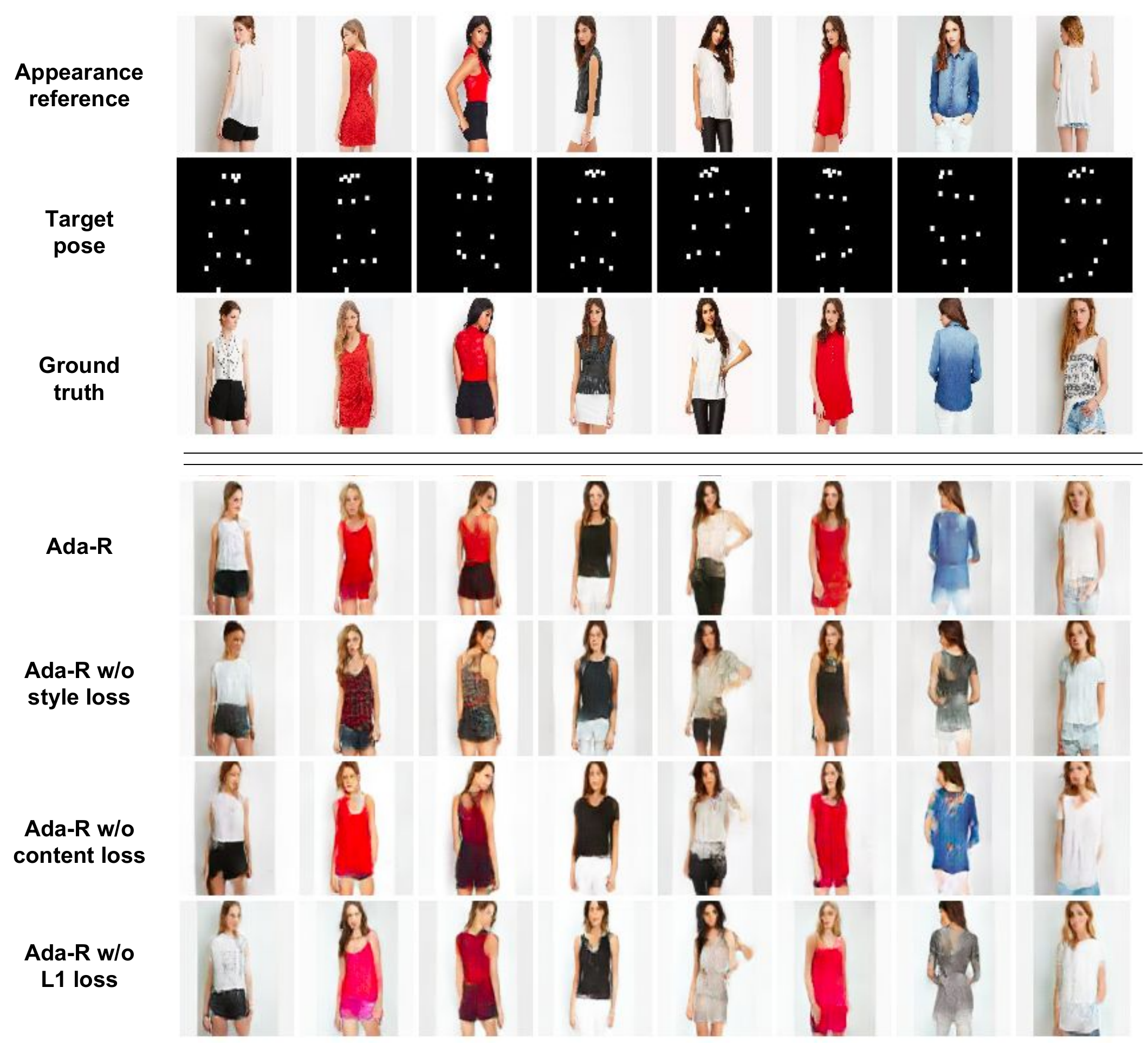}
\caption{Visualization of generation results of Ada-R using different loss terms on DeepFashion.}
\label{fig-fashion1}
\end{figure}

\begin{figure}[h]
\center
\includegraphics[width=0.95\textwidth]{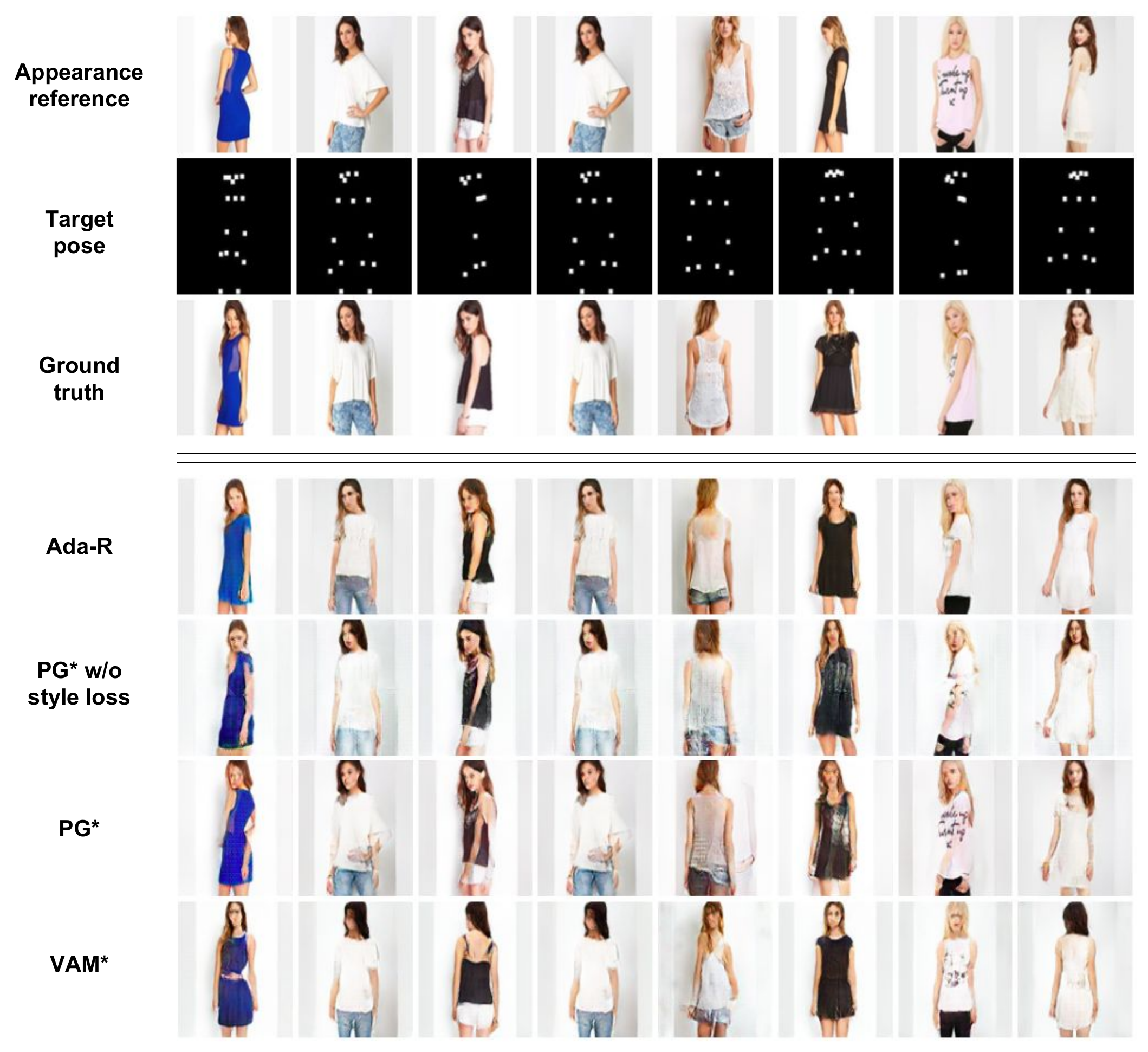}
\caption{Visualization of generation results of different state-of-the-art approaches on DeepFashion.}
\label{fig-fashion2}
\end{figure}

For evaluation, posemaps and images are normalized to $[-1,1]$. To measure the quality of the poses of the generated images, we propose a new \textit{perceptual pose score} which uses a state-of-the-art pose estimator~\cite{cao2017realtime} to extract pose from generated images and compares each pose joint with the corresponding ground truth pose joint using Euclidean distance. Quantitative results showing comparisons among our approach with all baselines are shown in Tab.~\ref{tbl:fashion_measure} and visualizations of all approaches are shown in Fig.~\ref{fig-fashion1} and Fig.~\ref{fig-fashion2}.

The quantitative evaluation results show that our \textit{Ada-R} models, except for the ablated one without $L_1$ loss, uniformly outperform \textit{Pose-GAN} and \textit{Visual Analogy Making} models by significant margins across all measures. The ablation study results for \textit{Ada-R} models on different components of the loss function also show that each component contributes positively to the model's performance. Since the target image directly supervises the model training through the content loss and $L_1$ loss, removing one of them, especially the $L_1$ loss, causes the largest degradation of model performance. This degradation in performance is also reflected in visualization results presented in Fig.~\ref{fig-fashion1}. As we can see from Fig.~\ref{fig-fashion1}, removing $L_1$ or content loss results in: (1) the pose of the target in the generated images being inferior (missing arms or hands); (2) images generated being more blurry; (3) the appearance of the generated images showing more errors in details (color transfer to arms, wrong colors, uneven colors in clothes). As seen from Tab.~\ref{tbl:fashion_measure}, removing the style loss from the training objective causes relatively smaller impacts on the model's performance and the model can generate person images striking the correct pose. However it is observed that many examples generated by \textit{Ada-R w/o style loss} do not reflect the colors of the style reference image correctly and the colors tend to be flat and unvaried compared with our full model.


When compared with \textit{PG$^\star$} models that learn the representation for the pose and the style together, \textit{Ada-R} which adopts a two-stream approach and encodes the pose and style separately also shows clear advantages. As shown in Fig.~\ref{fig-fashion2}, \textit{PG$^\star$} tends to generate images that show both the pose of the input posemap and the pose in the style reference image so as to match the color distribution of the style image, resulting in two overlapping targets in the generated images. While \textit{PG$^\star$ w/o style loss} tends to generate blurry images with incorrect poses or appearance as shown by the pose score and examples in Fig.~\ref{fig-fashion2}. The comparison demonstrates the advantage of disentangling the representations of pose and style and targeting each representation with a specific loss. \textit{Visual Analogy Making$^\star$ (VAM$^\star$)}~\cite{NIPS2015_5845_analogy}, also uses disentangled representations for pose and style and we use the same loss function as our model for training. However, we find that it cannot generate images with correct poses most of the time. Failure examples of \textit{Visual Analogy Making$^\star$} are presented in Fig. ~\ref{fig-fashion2}.

\begin{table}[h!]
  \centering
  \begin{tabular}{c  c  c  c  c  c  c  c }
  \hline
      & Ours & \makecell{Ours \\ w/o $\mathcal{L}_{s}$} & \makecell{Ours \\ w/o $\mathcal{L}_{C}$} & \makecell{Ours \\ w/o $L_1$} & PG & \makecell{PG \\ w/o $\mathcal{L}_{s}$} & VAM$^\star$
    \\ \hhline{========}
    MSE & \textbf{0.0849} & 0.0884 & 0.1070 & 0.1136 & 0.1276 & 0.1236 & 0.1652 
    \\ \hline
    PSNR & \textbf{17.2338} & 17.0346 & 16.1756 & 15.8898 & 15.5597 & 15.6150 & 14.3356
    \\ \hline
    SSIM & \textbf{0.6508} & 0.6484 & 0.6153 & 0.5810 & 0.5836 & 0.5708 & 0.5562
    \\ \hline
    Perceptual & \textbf{0.0310} & 0.0371 & 0.0671 & 0.0538 & 0.1046 & 0.0859 & 0.0910
    \\ \hline
    \end{tabular}
  \caption{Quantitative measures on DeepFashion.}\label{tbl:fashion_measure}
  \vspace{-0.5cm}
\end{table}

\subsection{Experiments on Volleyball}
\label{eval:volleyball}
\begin{figure*}[!tph]
\center
\includegraphics[width=0.9\textwidth]{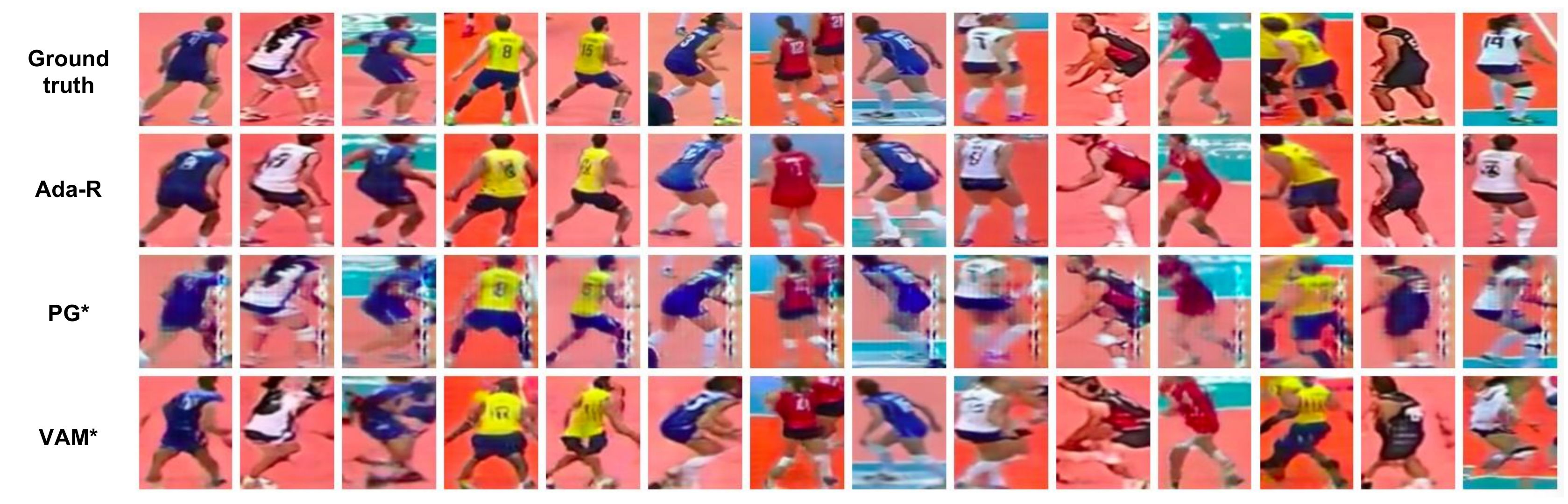}
\caption{Visualizations on Volleyball dataset.}
\label{fig:volley_visual}
\end{figure*}

The Volleyball dataset contains sequences of volleyball games. For this dataset, our model is trained to observe players in the $1^{st}$ input frames and predict their future appearances from frame 6 to frame 10. To get training sequences of each player, we run person detection~\cite{ren2015faster} and tracking~\cite{dlib09} to get tracklets of each player in each clip. We follow the data split of original dataset and preprocessing is conducted to filter out instances with less than 10 joints and clips containing less than 10 targets. We get 1262 clips for training and 790 clips for testing. Images of players are cropped and then resized to $256 \times 256$ pixels for both training and testing. For evaluations, we compare our model with state-of-the-art approaches including Pose-GAN and VAM as described in Sec.~\ref{eval:fashion}. Comparisons among our approach and two state-of-the-art approaches are shown in Tab.~\ref{tbl:volleyball_measure} and visualizations of all approaches are shown in Fig.~\ref{fig:volley_visual}. 

\begin{table}[h!]
  \centering
  \begin{tabular}{  c  c  c  c }
  \hline
      & Ours & PG w/o $\mathcal{L}_{s}$ & VAM
    \\ \hhline{====}
    MSE & \textbf{0.1670} & 0.1854 & 0.2091
    \\ \hline
    PSNR & \textbf{14.0191} & 13.5037 & 13.0174
    \\ \hline
    SSIM & \textbf{0.2825} & 0.2333 & 0.2178
    \\ \hline
    \end{tabular}
  \caption{Quantitative measures on Volleyball dataset.}\label{tbl:volleyball_measure}
\end{table}

As shown in both quantitative and qualitative results, our model performs very well on this application. Our model can generate clean and sharp person image sequences with correct appearance generated and consistent poses as in the posemap input, despite the complex scenario in terms of delicate pose changes, motion blur, and appearance variation.

\begin{figure*}[!tph]
\center
\includegraphics[width=0.9\textwidth]{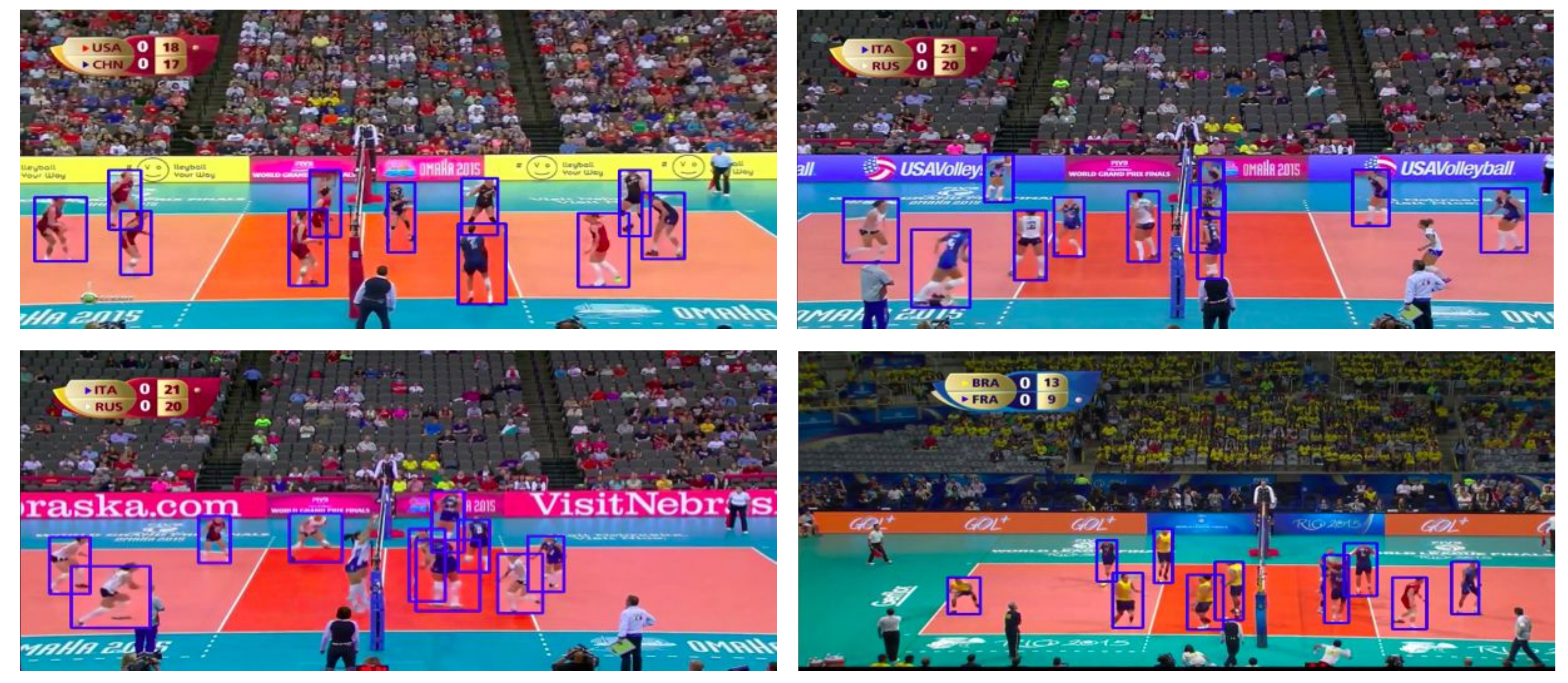}
\caption{Visualizations on whole frame generation on Volleyball dataset. The generated players are highlighted using \textcolor{blue}{blue} bounding boxes in each frame.}
\label{fig:volley_whole_frame}
\end{figure*}

We could also generate frames in original resolution $1280 \times 720$ by replacing players in a frame by generated ones. Examples of the frame synthesis results are shown in Fig.~\ref{fig:volley_whole_frame}, where the generated players are highlighted using blue bounding boxes.

\section{Conclusion}
We proposed a novel approach for synthesizing realistic images or sequences that generate the appearance of people taking the desired poses. Our approach encodes the appearance into convolutional filters. These filters are learned using a fully convolutional network, and utilized in an image-to-image translation structure that transfers the desired appearance to the desired pose. Both quantitative and qualitative results show that our model is superior to state-of-the-art approaches and can generate better images involving complex appearance and better videos involving complex human activities. The success of our model demonstrates that the combination of appearance filters and style loss can render the desired target appearance while adapting to the predicted pose.

\bibliography{egbib}

\begin{thebibliography}{33}
\providecommand{\natexlab}[1]{#1}
\providecommand{\url}[1]{\texttt{#1}}
\expandafter\ifx\csname urlstyle\endcsname\relax
  \providecommand{\doi}[1]{doi: #1}\else
  \providecommand{\doi}{doi: \begingroup \urlstyle{rm}\Url}\fi

\bibitem[Brand and Hertzmann(2000)]{BrandH00}
M.~Brand and A.~Hertzmann.
\newblock Style machines.
\newblock \emph{Conference on Special Interest Group on Computer Graphics and
  Interactive Techniques (SIGGRAPH)}, 2000.

\bibitem[Cao et~al.(2017)Cao, Simon, Wei, and Sheikh]{cao2017realtime}
Zhe Cao, Tomas Simon, Shih-En Wei, and Yaser Sheikh.
\newblock Realtime multi-person 2d pose estimation using part affinity fields.
\newblock In \emph{Conference on Computer Vision and Pattern Recognition
  (CVPR)}, 2017.

\bibitem[Chen et~al.(2017)Chen, Yuan, Liao, Yu, and Hua]{chen2017stylebank}
Dongdong Chen, Lu~Yuan, Jing Liao, Nenghai Yu, and Gang Hua.
\newblock Stylebank: An explicit representation for neural image style
  transfer.
\newblock In \emph{Conference on Computer Vision and Pattern Recognition
  (CVPR)}, 2017.

\bibitem[Efros et~al.(2003)Efros, Berg, Mori, and Malik]{EfrosBMM03}
A.A. Efros, A.C. Berg, G.~Mori, and J.~Malik.
\newblock Recognizing action at a distance.
\newblock In \emph{International Conference on Computer Vision (ICCV)}, 2003.

\bibitem[Finn et~al.(2016)Finn, Goodfellow, and
  Levine]{finn2016unsupervisedPhysicalInteraction}
Chelsea Finn, Ian Goodfellow, and Sergey Levine.
\newblock Unsupervised learning for physical interaction through video
  prediction.
\newblock In \emph{Advances in Neural Information Processing Systems (NIPS)},
  2016.

\bibitem[Gatys et~al.(2016)Gatys, Ecker, and
  Bethge]{gatys2015neuralStyleTransfer}
L.~A. Gatys, A.~S. Ecker, and M.~Bethge.
\newblock Image style transfer using convolutional neural networks.
\newblock In \emph{Conference on Computer Vision and Pattern Recognition
  (CVPR)}, 2016.

\bibitem[Goodfellow et~al.(2014)Goodfellow, Pouget-Abadie, Mirza, Xu,
  Warde-Farley, Ozair, Courville, and Bengio]{goodfellow2014GAN}
Ian Goodfellow, Jean Pouget-Abadie, Mehdi Mirza, Bing Xu, David Warde-Farley,
  Sherjil Ozair, Aaron Courville, and Yoshua Bengio.
\newblock Generative adversarial nets.
\newblock In \emph{Advances in Neural Information Processing Systems (NIPS)},
  2014.

\bibitem[Ibrahim et~al.(2016)Ibrahim, Muralidharan, Deng, Vahdat, and
  Mori]{Ibrahim_2016_CVPR_volleyball}
Mostafa~S. Ibrahim, Srikanth Muralidharan, Zhiwei Deng, Arash Vahdat, and Greg
  Mori.
\newblock A hierarchical deep temporal model for group activity recognition.
\newblock In \emph{Conference on Computer Vision and Pattern Recognition
  (CVPR)}, 2016.

\bibitem[Ioffe and Szegedy(2015)]{ioffe2015batch}
Sergey Ioffe and Christian Szegedy.
\newblock Batch normalization: Accelerating deep network training by reducing
  internal covariate shift.
\newblock In \emph{International Conference on Machine Learning (ICML)}, 2015.

\bibitem[Isola et~al.(2017)Isola, Zhu, Zhou, and Efros]{isola2016pix2pix}
Phillip Isola, Jun-Yan Zhu, Tinghui Zhou, and Alexei~A. Efros.
\newblock Image-to-image translation with conditional adversarial networks.
\newblock In \emph{Conference on Computer Vision and Pattern Recognition
  (CVPR)}, 2017.

\bibitem[Johnson et~al.(2016)Johnson, Alahi, and
  Fei-Fei]{johnson2016perceptualLoss}
Justin Johnson, Alexandre Alahi, and Li~Fei-Fei.
\newblock Perceptual losses for real-time style transfer and super-resolution.
\newblock In \emph{European Conference on Computer Vision (ECCV)}, 2016.

\bibitem[Karacan et~al.(2016)Karacan, Akata, Erdem, and Erdem]{KaracanAEE16}
Levent Karacan, Zeynep Akata, Aykut Erdem, and Erkut Erdem.
\newblock Learning to generate images of outdoor scenes from attributes and
  semantic layouts.
\newblock \emph{CoRR}, 2016.

\bibitem[King(2009)]{dlib09}
Davis~E. King.
\newblock Dlib-ml: A machine learning toolkit.
\newblock \emph{Journal of Machine Learning Research(JMLR)}, 10:\penalty0
  1755--1758, 2009.

\bibitem[Liu et~al.(2016)Liu, Luo, Qiu, Wang, and
  Tang]{liuLQWTcvpr16DeepFashion}
Ziwei Liu, Ping Luo, Shi Qiu, Xiaogang Wang, and Xiaoou Tang.
\newblock Deepfashion: Powering robust clothes recognition and retrieval with
  rich annotations.
\newblock In \emph{Proceedings of IEEE Conference on Computer Vision and
  Pattern Recognition (CVPR)}, 2016.

\bibitem[Liu et~al.(2017)Liu, Yeh, Tang, Liu, and Agarwala]{liu2017video}
Ziwei Liu, Raymond~A. Yeh, Xiaoou Tang, Yiming Liu, and Aseem Agarwala.
\newblock Video frame synthesis using deep voxel flow.
\newblock In \emph{International Conference on Computer Vision (ICCV)}, 2017.

\bibitem[Ma et~al.(2017{\natexlab{a}})Ma, Jia, Sun, Schiele, Tuytelaars, and
  Van~Gool]{ma2017pose}
Liqian Ma, Xu~Jia, Qianru Sun, Bernt Schiele, Tinne Tuytelaars, and Luc
  Van~Gool.
\newblock Pose guided person image generation.
\newblock In \emph{Advances in Neural Information Processing Systems}, pages
  406--416, 2017{\natexlab{a}}.

\bibitem[Ma et~al.(2017{\natexlab{b}})Ma, Sun, Georgoulis, Gool, Schiele, and
  Fritz]{MaDPIG17}
Liqian Ma, Qianru Sun, Stamatios Georgoulis, Luc~Van Gool, Bernt Schiele, and
  Mario Fritz.
\newblock Disentangled person image generation.
\newblock \emph{CoRR}, 2017{\natexlab{b}}.

\bibitem[Mansimov et~al.(2016)Mansimov, Parisotto, Ba, and
  Salakhutdinov]{mansimov2015generating}
Elman Mansimov, Emilio Parisotto, Jimmy~Lei Ba, and Ruslan Salakhutdinov.
\newblock Generating images from captions with attention.
\newblock \emph{International Conference on Learning Representations (ICLR)},
  2016.

\bibitem[Mathieu et~al.(2016)Mathieu, Couprie, and
  LeCun]{mathieu2015deepBeyondMSE}
Michael Mathieu, Camille Couprie, and Yann LeCun.
\newblock Deep multi-scale video prediction beyond mean square error.
\newblock In \emph{International Conference on Learning Representations
  (ICLR)}, 2016.

\bibitem[Oh et~al.(2015)Oh, Guo, Lee, Lewis, and
  Singh]{oh2015actionConditionedVideoPrediction}
Junhyuk Oh, Xiaoxiao Guo, Honglak Lee, Richard~L Lewis, and Satinder Singh.
\newblock Action-conditional video prediction using deep networks in atari
  games.
\newblock In \emph{Advances in Neural Information Processing Systems (NIPS)},
  2015.

\bibitem[Ranzato et~al.(2014)Ranzato, Szlam, Bruna, Mathieu, Collobert, and
  Chopra]{ranzato2014videoGenerationBaseline}
MarcAurelio Ranzato, Arthur Szlam, Joan Bruna, Michael Mathieu, Ronan
  Collobert, and Sumit Chopra.
\newblock Video (language) modeling: a baseline for generative models of
  natural videos.
\newblock \emph{arXiv preprint arXiv:1412.6604}, 2014.

\bibitem[Reed et~al.(2016)Reed, Akata, Yan, Logeswaran, Schiele, and
  Lee]{reed2016generative}
Scott Reed, Zeynep Akata, Xinchen Yan, Lajanugen Logeswaran, Bernt Schiele, and
  Honglak Lee.
\newblock Generative adversarial text to image synthesis.
\newblock In \emph{International Conference on Machine Learning (ICML)}, 2016.

\bibitem[Reed et~al.(2015)Reed, Zhang, Zhang, and Lee]{NIPS2015_5845_analogy}
Scott~E Reed, Yi~Zhang, Yuting Zhang, and Honglak Lee.
\newblock Deep visual analogy-making.
\newblock In \emph{Advances in Neural Information Processing Systems (NIPS)},
  2015.

\bibitem[Ren et~al.(2015)Ren, He, Girshick, and Sun]{ren2015faster}
Shaoqing Ren, Kaiming He, Ross Girshick, and Jian Sun.
\newblock Faster r-cnn: Towards real-time object detection with region proposal
  networks.
\newblock In \emph{Advances in Neural Information Processing Systems (NIPS)},
  2015.

\bibitem[Sadeghi et~al.(2015)Sadeghi, Zitnick, and
  Farhadi]{sadeghi2015visalogy}
Fereshteh Sadeghi, C~Lawrence Zitnick, and Ali Farhadi.
\newblock Visalogy: Answering visual analogy questions.
\newblock In \emph{Advances in Neural Information Processing Systems (NIPS)},
  2015.

\bibitem[Simonyan and Zisserman(2015)]{Simonyan15VGG}
K.~Simonyan and A.~Zisserman.
\newblock Very deep convolutional networks for large-scale image recognition.
\newblock In \emph{International Conference on Learning Representations
  (ICLR)}, 2015.

\bibitem[Srivastava et~al.(2015)Srivastava, Mansimov, and
  Salakhudinov]{srivastava2015unsupervisedVideoRepresentationLSTMs}
Nitish Srivastava, Elman Mansimov, and Ruslan Salakhudinov.
\newblock Unsupervised learning of video representations using lstms.
\newblock In \emph{International Conference on Machine Learning (ICML)}, 2015.

\bibitem[Villegas et~al.(2017)Villegas, Yang, Zou, Sohn, Lin, and
  Lee]{Villegas2017LearningToGenerate}
Ruben Villegas, Jimei Yang, Yuliang Zou, Sungryull Sohn, Xunyu Lin, and Honglak
  Lee.
\newblock Learning to generate long-term future via hierarchical prediction.
\newblock In \emph{International Conference on Machine Learning (ICML)}, 2017.

\bibitem[Walker et~al.(2017)Walker, Marino, Gupta, and
  Hebert]{walker2017poseKnow}
Jacob Walker, Kenneth Marino, Abhinav Gupta, and Martial Hebert.
\newblock The pose knows: Video forecasting by generating pose futures.
\newblock In \emph{International Conference on Computer Vision (ICCV)}, 2017.

\bibitem[Wang et~al.(2004)Wang, Bovik, Sheikh, and Simoncelli]{wang2004image}
Zhou Wang, Alan~C Bovik, Hamid~R Sheikh, and Eero~P Simoncelli.
\newblock Image quality assessment: from error visibility to structural
  similarity.
\newblock \emph{IEEE transactions on image processing}, 13\penalty0
  (4):\penalty0 600--612, 2004.

\bibitem[Zhao et~al.(2017)Zhao, Wu, Cheng, Liu, and Feng]{ZhaoWCLF17}
Bo~Zhao, Xiao Wu, Zhi{-}Qi Cheng, Hao Liu, and Jiashi Feng.
\newblock Multi-view image generation from a single-view.
\newblock \emph{CoRR}, abs/1704.04886, 2017.

\bibitem[Zhu et~al.(2017{\natexlab{a}})Zhu, Park, Isola, and
  Efros]{CycleGAN2017}
Jun-Yan Zhu, Taesung Park, Phillip Isola, and Alexei~A. Efros.
\newblock Unpaired image-to-image translation using cycle-consistent
  adversarial networks.
\newblock In \emph{International Conference on Computer Vision (ICCV)},
  2017{\natexlab{a}}.

\bibitem[Zhu et~al.(2017{\natexlab{b}})Zhu, Fidler, Urtasun, Lin, and
  Loy]{ZhuPrada17}
Shizhan Zhu, Sanja Fidler, Raquel Urtasun, Dahua Lin, and Chen~Change Loy.
\newblock Be your own prada: Fashion synthesis with structural coherence.
\newblock \emph{CoRR}, 2017{\natexlab{b}}.

\end{thebibliography}
\end{document}